\documentclass[letterpaper, 10 pt, conference]{ieeeconf}  

\makeatletter
\let\NAT@parse\undefined
\makeatother

\IEEEoverridecommandlockouts                              

\usepackage{graphicx} 
\usepackage[numbers]{natbib}
\usepackage{amsmath} 
\usepackage{amssymb}  
\usepackage{mathtools}
\usepackage[]{algorithm2e}
\usepackage[caption=false,font=footnotesize]{subfig}

\newcommand{\figref}[1]{Fig.~\ref{#1}}
\newcommand{\secref}[1]{Section~\ref{#1}}

\title{\LARGE \bf
Project AutoVision: Localization and 3D Scene Perception for an Autonomous Vehicle with a Multi-Camera System
}

\author{
\authorblockN{Lionel Heng$^{1}$, Benjamin Choi$^{1}$, Zhaopeng Cui$^{2}$, Marcel Geppert$^{2}$, Sixing Hu$^{3}$, Benson Kuan$^{1}$, Peidong Liu$^{2}$,}
\authorblockN{Rang Nguyen$^{3}$, Ye Chuan Yeo$^{1}$, Andreas Geiger$^{4}$, Gim Hee Lee$^{3}$, Marc Pollefeys$^{2,5}$, and Torsten Sattler$^{6}$}
\thanks{$^{1}$DSO National Laboratories}%
\thanks{$^{2}$ETH Z\"{u}rich}%
\thanks{$^{3}$National University of Singapore}%
\thanks{$^{4}$MPI-IS and University of T\"{u}bingen}%
\thanks{$^{5}$Microsoft, Switzerland}%
\thanks{$^{6}$Chalmers University of Technology, Sweden}%
}

\begin{document}

\maketitle
\thispagestyle{empty}
\pagestyle{empty}

\begin{abstract}
Project AutoVision aims to develop localization and 3D scene perception capabilities for a self-driving vehicle. Such capabilities will enable autonomous navigation in urban and rural environments, in day and night, and with cameras as the only exteroceptive sensors. The sensor suite employs many cameras for both 360-degree coverage and accurate multi-view stereo; the use of low-cost cameras keeps the cost of this sensor suite to a minimum. In addition, the project seeks to extend the operating envelope to include GNSS-less conditions which are typical for environments with tall buildings, foliage, and tunnels. Emphasis is placed on leveraging multi-view geometry and deep learning to enable the vehicle to localize and perceive in 3D space. This paper presents an overview of the project, and describes the sensor suite and current progress in the areas of calibration, localization, and perception.
\end{abstract}

\section{INTRODUCTION}

The three DARPA Grand Challenges in the last decade set off a wave of disruption in the automotive industry. With widespread belief that autonomous vehicles can revolutionize logistics and mobility, automakers and technology companies are racing with one another to put autonomous vehicles on the road within the next few years. LiDAR sensors are the primary sensing modality for a vast majority of autonomous vehicles; they generate highly accurate 3D point cloud data in both day and night, and enable localization and 3D scene perception at all times of the day. In contrast, cameras require sufficient ambient lighting, and do not directly provide 3D point cloud data. However, cameras yield high-resolution image data which better facilitates scene segmentation and understanding. In addition, we can leverage multi-view geometry techniques to infer depth data from multiple cameras albeit with lower accuracy than depth data from LiDAR sensors. Cameras can be fitted with either wide-field-of-view or fisheye lenses, giving them a significantly larger vertical field of view and higher vertical resolution compared to LiDAR sensors. In Project AutoVision, we choose to focus on cameras as the sole sensing modality for autonomous vehicles; we observe that much research remains to be done in realising robust visual localization and perception for autonomous vehicles.

Project AutoVision started in late 2016 with the goal to develop localization and 3D scene perception algorithms for autonomous vehicles exclusively equipped with cameras. Project AutoVision is similar to, but differs from Project V-Charge \citep{Furgale2013IV, Schwesinger2016IV, Haene2017IVC} in the aspect that Project AutoVision extends the operating envelope from parking lots and garages to large-scale urban and rural environments with higher driving speeds and widely varying illumination conditions. Parallels can also be drawn between Project AutoVision and AutoX, both of which only rely on cameras for localization and perception. However, little is known about AutoX's localization and perception approaches due to commercial interests.

We aim to localize in both mapped and unmapped areas without relying on GNSS; we do not want to limit the vehicle's operation to mapped areas in GNSS-less conditions. In addition, we work towards real-time 3D mapping as a 3D geometric map can aid navigation of multi-level structures and higher-level scene perception tasks such as terrain analysis and 3D scene understanding.
We follow the traditional approach \citep{Ziegler2014ITS, Broggi2015ITS} of applying multi-view geometry to localization and 3D geometric mapping, and machine learning to cross-modal matching, object detection, and scene segmentation. On the other hand, with the advent of deep neural networks, people \citep{bojarski2016CoRR} have used end-to-end learning for vision-based autonomous vehicles but with limited success.

We make the following contributions:
\begin{enumerate}
\item Real-time visual-inertial odometry with a multi-camera system \citep{Liu2017IROS, Liu2018IROS}.
\item Real-time GNSS-less visual localization in unmapped areas using geo-referenced satellite imagery and without GNSS, assuming that the initial global position and heading of the vehicle are known \citep{Hu2018CVPR, Hu2019IJCV}.
\item Real-time GNSS-less visual localization with a multi-camera system in mapped areas using a geo-referenced sparse 3D map and without prior knowledge of the vehicle's global pose \citep{Geppert2019ICRA}.
\item Real-time 3D dense mapping with a multi-fisheye-camera system \citep{Cui2019ICRA}.
\end{enumerate}
In this paper, we briefly describe these contributions which have been integrated into a single working system. The reader can refer to our published work for more details on the algorithms and experimental results. In addition, the paper gives details of the hardware setup, the software architecture, and the automated methods used for calibrating the multi-sensor suite; such details are not found in our published work on individual localization and perception components.

\section{System}

In this section, we give an overview of the sensors on the AutoVision vehicle platform, and the software architecture that enables various software modules to work together to enable the vehicle to localize and perceive in 3D.

\begin{figure}
  \centering
  \includegraphics[width=0.95\columnwidth]{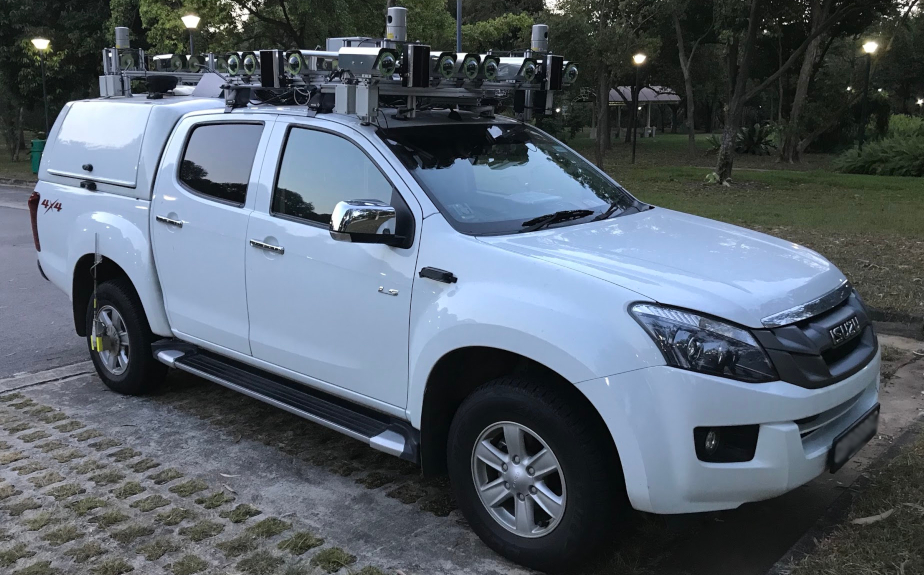}
  \caption{The AutoVision vehicle platform.}
  \label{fig:autovision_vehicle}
\end{figure}

\begin{figure}
  \centering
  \includegraphics[width=0.95\columnwidth]{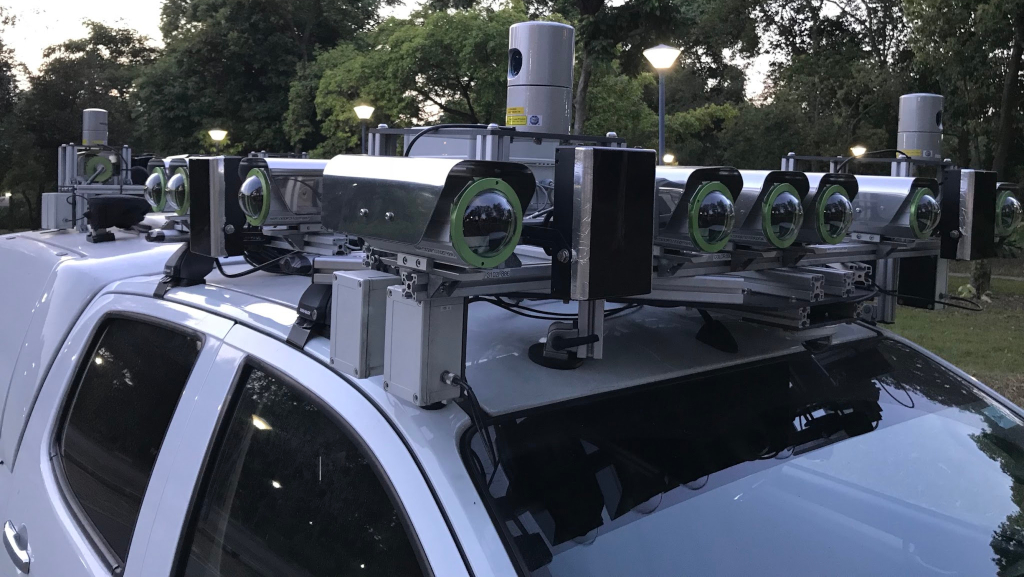}
  \caption{A close-up of the sensor suite on the AutoVision vehicle platform.}
  \label{fig:autovision_sensor_suite}
\end{figure}

\begin{figure}
  \includegraphics[width=\columnwidth,trim={0 5mm 0 0},clip]{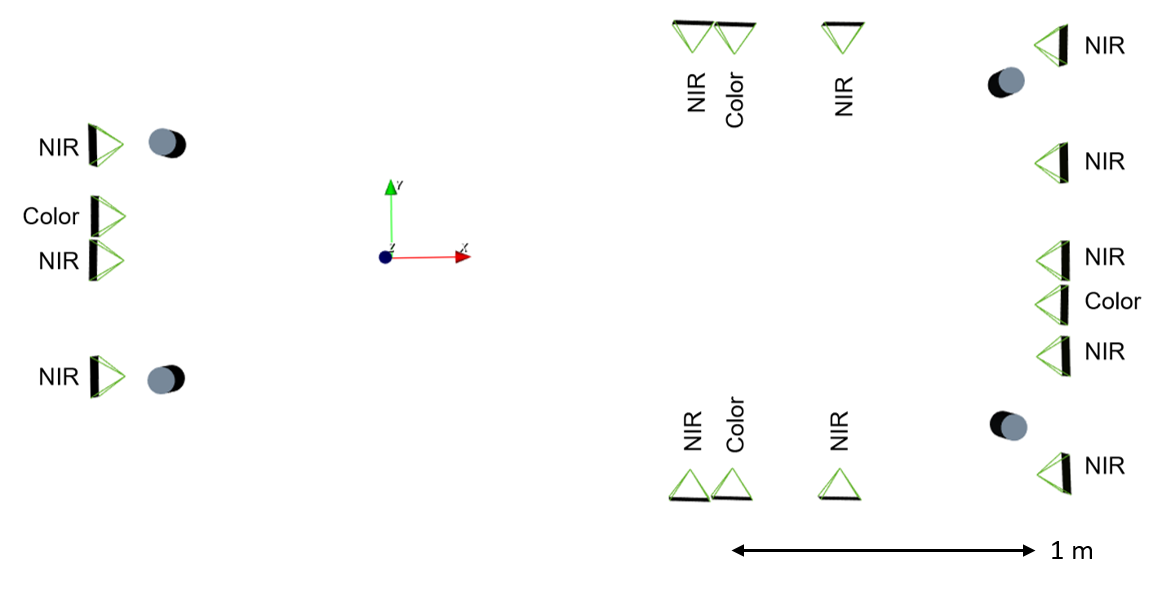}
  \caption{The sensor layout on the AutoVision vehicle platform. Each green-colored frustum and grey-colored cylinder represent a camera and LiDAR sensor respectively. The origin of the three axes indicates the location of the IMU. The red-colored $x$-axis points towards the front of the vehicle while the green-colored $y$-axis points towards the left of the vehicle. The extrinsic transformation between each sensor and the IMU was estimated with automated calibration tools.}
  \label{fig:autovision_sensor_system}
\end{figure}

\begin{figure*}
    \centering
  \subfloat[Day-time.]{%
       \includegraphics[width=0.32\textwidth]{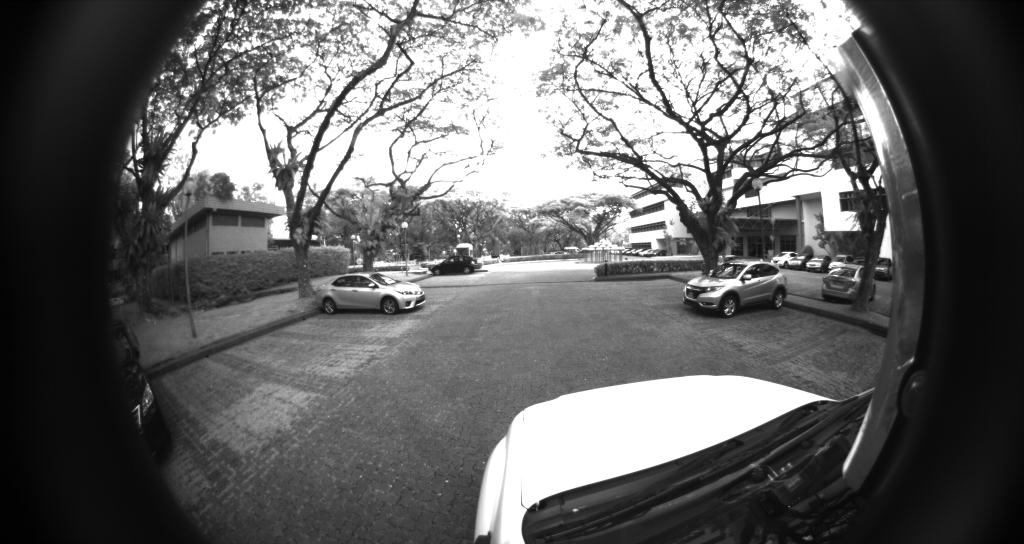}}
    \label{fig:image_day}\hfill
  \subfloat[Night-time with NIR illumination.]{%
        \includegraphics[width=0.32\textwidth]{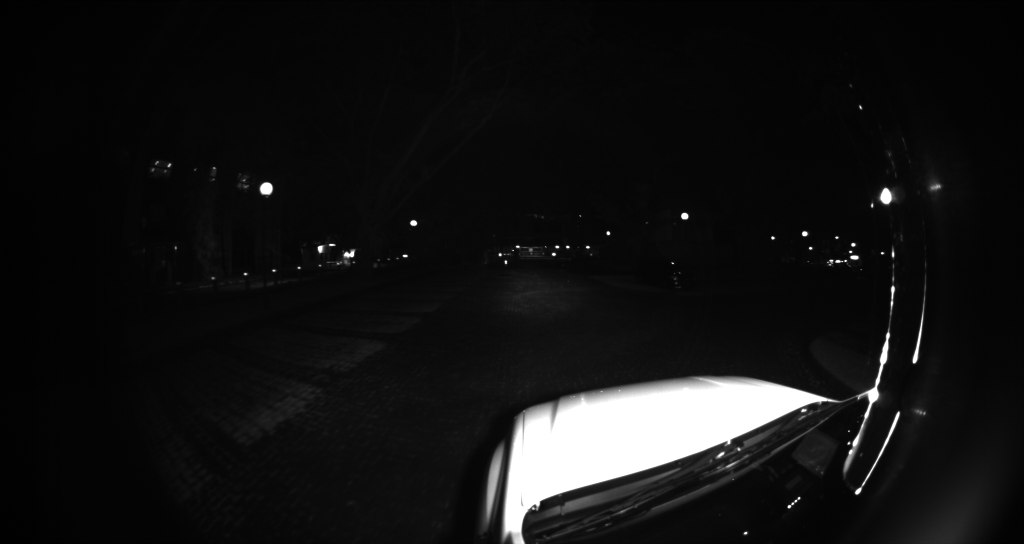}}
    \label{1c}\hfill
  \subfloat[Night-time without NIR illumination.]{%
        \includegraphics[width=0.32\textwidth]{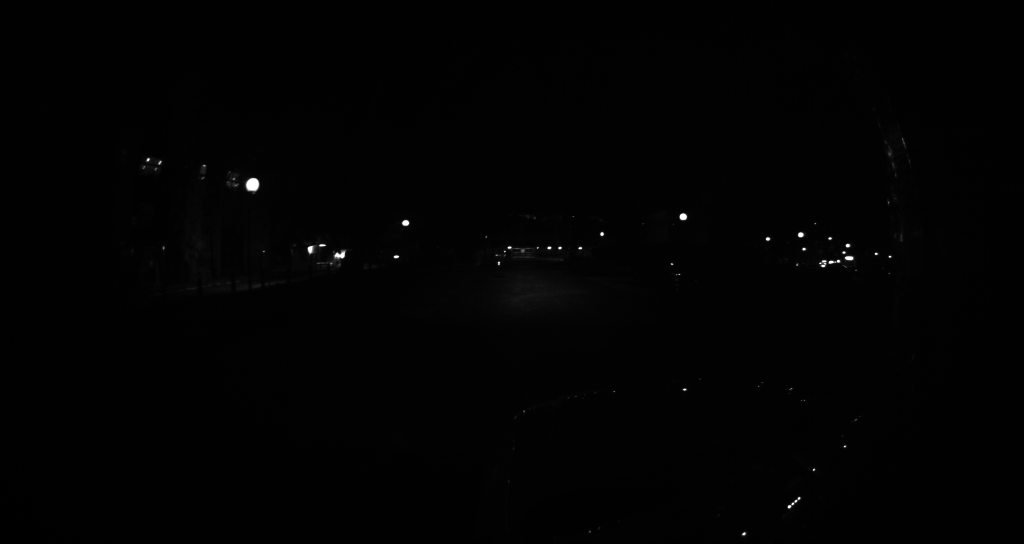}}
     \label{fig:image_night_no_illum} 
  \caption{Images captured from a front left camera on the AutoVision vehicle and in varying lighting conditions.}
  \label{fig:images_lighting} 
\end{figure*}

Our AutoVision vehicle platform is a Isuzu D-Max pick-up truck which has been modified to include a drive-by-wire system for autonomous driving. \figref{fig:autovision_vehicle} shows the vehicle platform while \figref{fig:autovision_sensor_suite} shows a close-up view of the sensors on the vehicle roof. Four color cameras and twelve NIR cameras are fitted with $180^\circ$-field-of-view fisheye lenses and installed in a surround-view configuration on top of the vehicle. All cameras output 2-megapixel images at 30 Hz, and are set to automatic exposure mode so that they can adapt to changing lighting conditions. We only use 12-bit grayscale images from the NIR cameras as input to all localization and perception modules; color cameras are only used for visualization purposes. NIR cameras are more light-sensitive than color cameras, and can detect light in both the visible and NIR wavelengths. In addition, NIR cameras provide sharp and clean images unlike Bayer-encoded color images that suffer from demosaicing artefacts. We use NIR cameras in conjunction with NIR illuminators which can improve low-light imaging quality and whose illumination is invisible to and does not distract drivers on the road. \figref{fig:images_lighting} shows examples of images captured at approximately the same location in varying lighting conditions. Camera enclosures provide cameras with IP67 protection from the weather elements. \figref{fig:autovision_sensor_system} shows the camera layout on the vehicle roof. The front side of the vehicle has the highest number of NIR cameras; these 5 NIR cameras facilitate wide-baseline stereo, and in turn, long-range perception which is critical for autonomous vehicles moving forward at high speeds. We exploit the dominantly longitudinal movement of the vehicle by simulating a multi-baseline stereo system on each side of the vehicle and which consists of 2 actual NIR cameras and at least 1 virtual NIR camera.

\subsection{Hardware}

A dual-antenna GNSS/INS system with a tactical-grade IMU is installed in the vehicle. Data from this GNSS/INS system is post-processed offline to yield centimeter-level ground-truth position and attitude data which is used to evaluate localization accuracy. A 3D LiDAR sensor is mounted at each of the four corners of the vehicle roof; fused point cloud data from all 4 LiDAR sensors is used to evaluate perception accuracy. All sensor data is hardware-timestamped to sub-microsecond precision. Such accurate time synchronization is made possible through the use of a time server. This time server synchronizes to the GNSS/INS system via PPS signals and NMEA data. In turn, LiDAR sensors and cameras time-synchronize to the time server via PPS/NMEA and PTP respectively. All sensors interface over a 10 GbE network switch with multiple industrial-grade computers equipped with GPUs. The 16-camera system yields 1.38 GB/s of image data which we record simultaneously to 4 solid-state drives (SSDs); we configure each computer to have two SSDs, and thus, two computers are required for data recording. 

\subsection{Software}
\begin{figure*}
  \centering
  \includegraphics[width=0.8\textwidth]{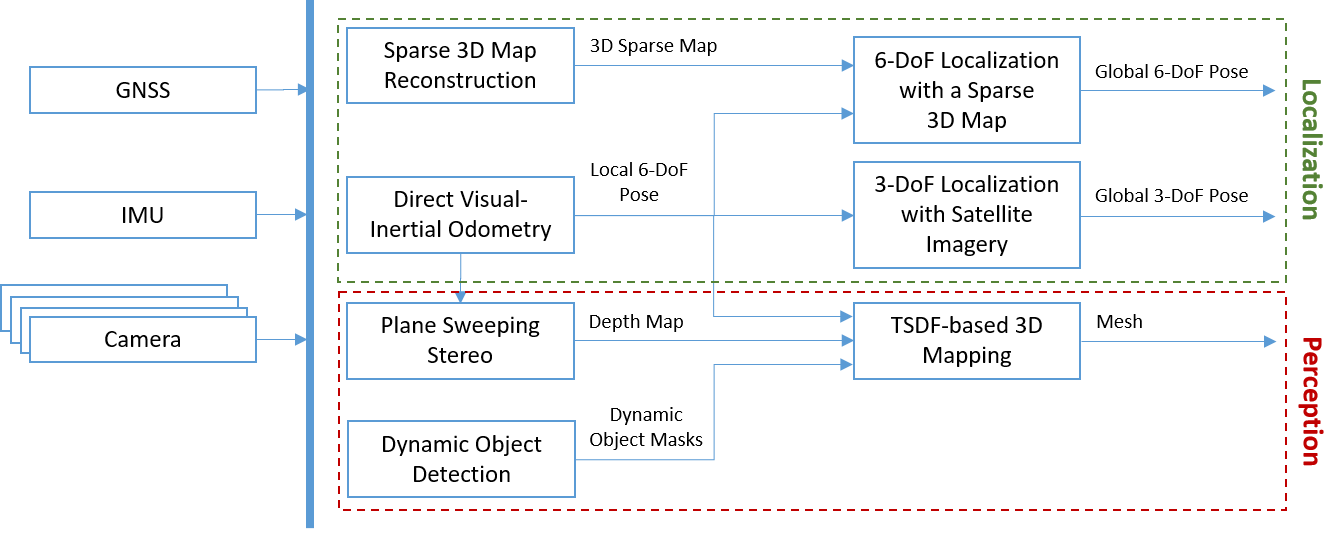}
  \caption{Our AutoVision software architecture. GNSS is only used for sparse 3D map reconstruction.}
  \label{fig:software_architecture}
\end{figure*}

Our software stack is based on the ROS 2 software framework, and runs on Windows 10. We use RTI Connext DDS for inter-process communications. \figref{fig:software_architecture} shows our software architecture; with the exception of the block representing sparse 3D map reconstruction which runs offline, each block represents a node which subscribes and publishes to topics. GNSS and IMU measurements are only used by sparse 3D map reconstruction and visual-inertial odometry respectively. All nodes subscribe to image data.

\section{Calibration}
An accurate calibration is an essential prerequisite for localization and perception to work well with a multi-camera system. Our calibration pipeline is automated and involves the following steps in order: intrinsic and extrinsic calibration of a multi-camera system, extrinsic calibration between a calibrated multi-camera system and a GNSS/INS system, and extrinsic calibration between a LiDAR sensor and a calibrated multi-camera system.

\begin{figure}
  \centering
  \includegraphics[width=0.9\columnwidth]{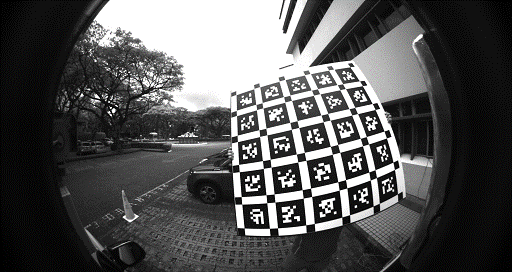}
  \caption{The fiducial target used for calibration of the multi-sensor system.}
  \label{fig:camera_calibration}
\end{figure}

We do intrinsic and extrinsic calibration of the multi-camera system with the help of a fiducial target which is a grid of AprilTag markers \citep{Wang2016IROS} with known dimensions, and is shown in \figref{fig:camera_calibration}. As each AprilTag marker has an unique identifier, calibration will work even if multiple cameras observe different parts of the target. This versatility comes in handy when calibrating pairs of cameras at the vehicle's corners and with perpendicular optical axes; it is difficult for such a pair of cameras to observe the entire target. For the intrinsic calibration, we can choose from multiple camera projection and distortion models. In this project, we use the unified projection model \citep{Geyer2000ECCV,Barreto2001CVPR} and the plumb bob distortion model \citep{Brown1966PE}. We also perform a photometric calibration of each camera using the method described in \citep{Engel2016CoRR}. This photometric calibration is useful for photometric-based matching between images captured with different exposure times, for example, in direct visual odometry and plane-sweeping stereo.

We obtain the extrinsic transformation between the calibrated multi-camera system and the GNSS/INS system by following an approach similar to that of \citet{Heng2015JFR, Heng2015AURO}. Here, the reference frame of the GNSS/INS system coincides with that of the IMU. We run semi-direct stereo visual odometry (VO) \citep{Heng2016IROS} for a stereo pair on each side of the vehicle. Each instance of stereo VO yields a set of camera poses and feature tracks. From hand-eye calibration using the reference camera's poses from stereo VO and the GNSS/INS system's poses, we obtain an initial estimate of the extrinsic transformation. Subsequently, we refine the extrinsic transformation by solving a non-linear least-squares problem in which we minimize the sum of squared reprojection errors associated with feature tracks while keeping the GNSS/INS system's poses and inter-camera transformations fixed.

\begin{figure}
  \centering
  \includegraphics[width=0.9\columnwidth]{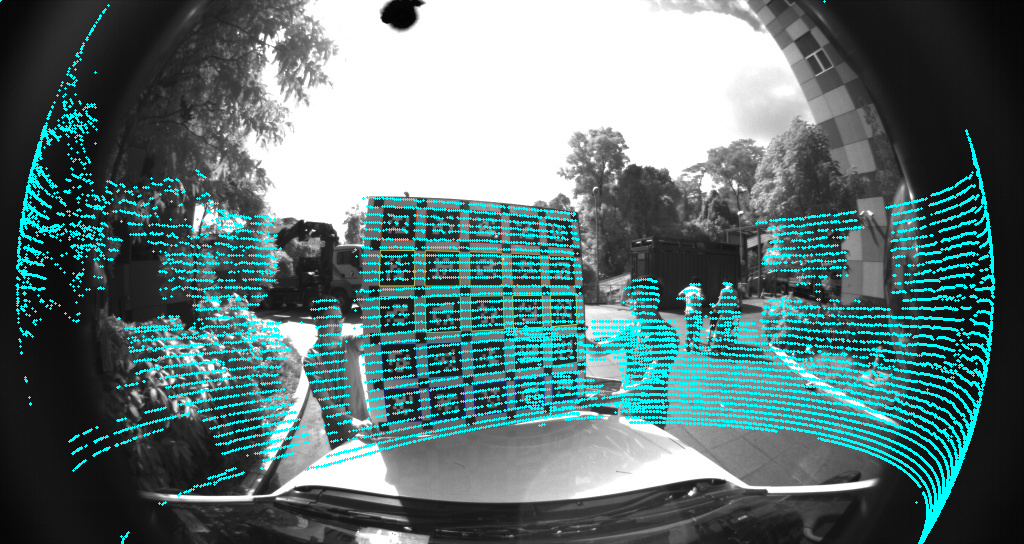}
  \caption{Projeciton of LiDAR scan points into a camera image.}
  \label{fig:lidar_camera_system_calibration}
\end{figure}

With the same fiducial target from intrinsic and extrinsic calibration of the multi-camera system, we perform extrinsic calibration between each LiDAR sensor and the calibrated multi-camera system. Given a set of images captured simultaneously from the multi-camera system, we detect the fiducial target in each image, and estimate its pose with respect to the multi-camera system by minimizing the squared sum of reprojection errors across all images in which the target was detected. At the same time, we identify the set of points corresponding to the fiducial target in the LiDAR scan by using plane segmentation \citep{Geiger2012ICRA}, and estimate the target's plane parameters with respect to the LiDAR sensor. With repeated observations of the target in different orientations, we independently estimate the rotation and translation components of the extrinsic transformation between the LiDAR sensor and the multi-camera system by doing singular value decomposition and solving a linear system of equations respectively. Subsequently, we refine the extrinsic transformation by minimizing the sum of squared point-plane errors; the points form part of the LiDAR scan identified as corresponding to the target, and the plane parameters are inferred from the estimated pose of the target with respect to the multi-camera system. \figref{fig:lidar_camera_system_calibration} shows the projection of LiDAR scan points into a camera image using the results of extrinsic calibration between the LiDAR sensor and the multi-camera system.

\section{Localization}

One goal of Project AutoVision is to enable an autonomous vehicle to localize in both unmapped and premapped environments. A vehicle relying on map-based localization is restricted to movement within the map. We want to allow the vehicle to navigate beyond the map into unmapped areas by leveraging satellite imagery. However, a premapped environment enables the vehicle to localize with higher accuracy. Global pose estimates are susceptible to pose jumps; smooth local pose estimates are required for stable path tracking and to build consistent 3D maps. For this purpose, we use direct visual-inertial odometry which runs at the frame rate of the multi-camera system.

\subsection{Direct Visual-Inertial Odometry}

\begin{figure}
  \includegraphics[width=0.9\columnwidth]{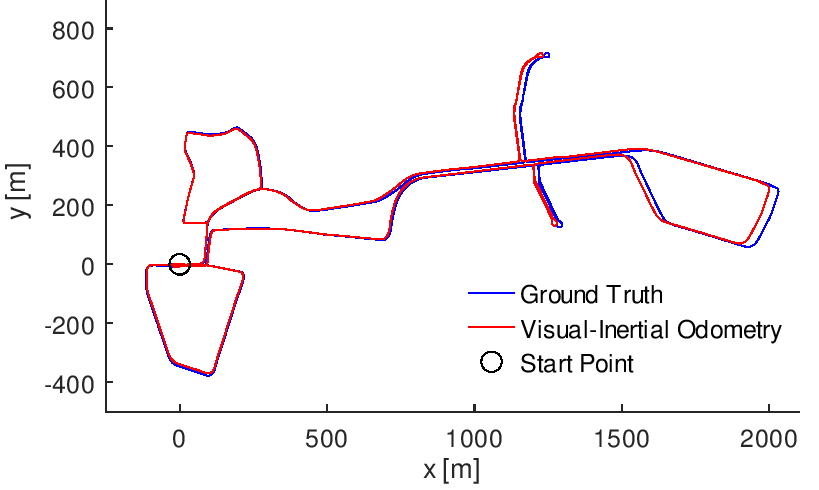}
  \caption{The positions estimated by VIO vs ground truth positions. One front stereo pair and one rear stereo pair were used for VIO.}
  \label{fig:vio}
\end{figure}

Our direct visual-inertial odometry (VIO) implementation for a multi-camera system \citep{Liu2017IROS, Liu2018IROS} estimates the local pose of the vehicle at 30 Hz. Our direct VIO implementation contains two threads: (1) the tracking thread estimates the local pose by minimizing photometric errors between the most recent keyframe and the current frame, and (2) the mapping thread initializes the depth of all sampled feature points using plane-sweeping stereo, and uses a sliding window optimizer to refine poses and structure jointly. Extensive experiments described by \citet{Liu2018IROS} show our implementation to work robustly for a 4-stereo-camera configuration with less than 1\% translational drift in day-time and night-time with NIR illumination, and less than 2\% translational drift in night-time without NIR illumination. \figref{fig:vio} plots the pose estimates from our VIO implementation against ground truth data for a 8.2km route in a route covering both urban and rural environments.

\subsection{3-DoF Localization with Satellite Imagery}
\label{sec:sat_map_localization}

In areas that have not been premapped, we rely on satellite imagery to estimate the 3-DoF global pose. Specifically, we estimate the $(x,y)$ position and heading of the vehicle with respect to the UTM coordinate frame. Assuming that the initial position and heading are known from user input, we use the particle filter approach in which we use local pose data from VIO for particle propagation and output from a deep network \citep{Hu2018CVPR, Hu2019IJCV} for particle weighting.

\begin{figure}
  \centering
  \includegraphics[width=0.9\columnwidth]{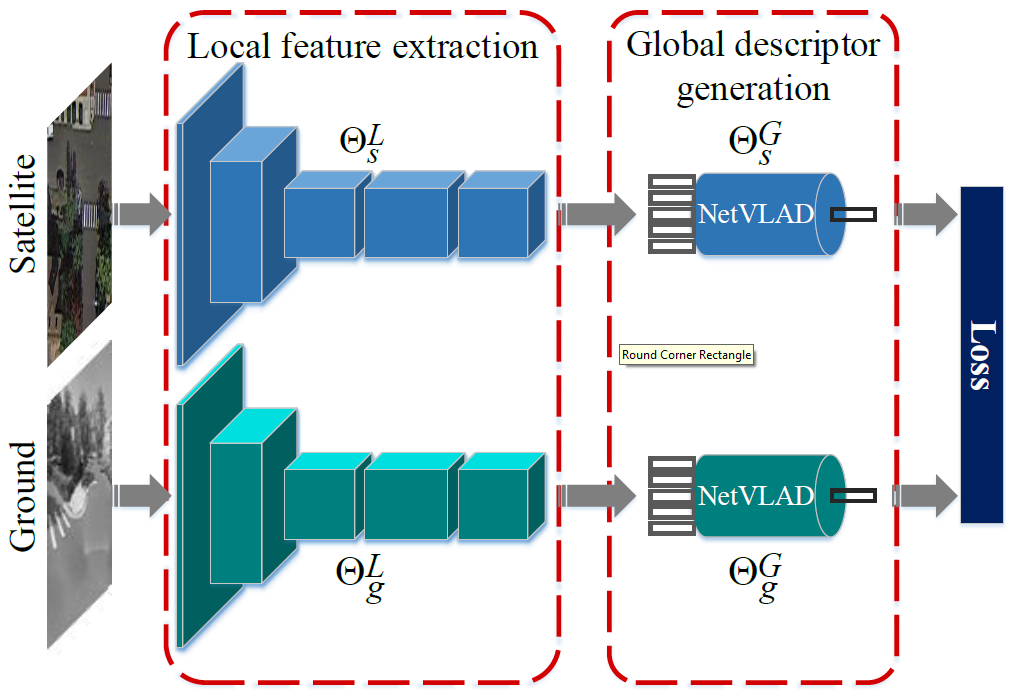}
  \caption{The architecture of our deep network, CVM-Net, for cross-view matching \citep{Hu2018CVPR, Hu2019IJCV}.}
  \label{fig:cvm_net}
\end{figure}

As shown in \figref{fig:cvm_net}, the deep network called CVM-Net is a Siamese network that takes satellite and ground-level panoramic images as input. We obtain the panoramic image by stitching the cylindrical projections of images taken from four cameras: one camera on each side of the vehicle. For each image, we extract local features via fully convolutional networks. Two aligned NetVLADs aggregate local features from both images into global descriptors that are in a common space for similarity comparison. The weight for each particle is inversely proportional to the Euclidean distance between the global descriptors corresponding to the ground-level panoramic image and the satellite image patch nearest the particle's position.

\begin{figure}
  \includegraphics[width=\columnwidth]{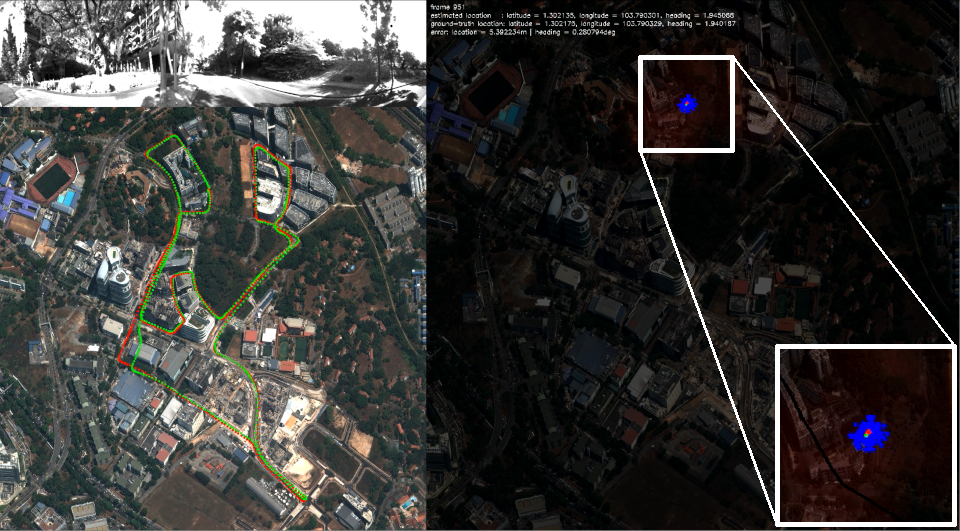}
  \caption{A visualization of our satellite-imagery-based localisation in an urban environment.}
  \label{fig:sat_map_loc}
\end{figure}

We run two experiments with a 5km route in both an urban environment and a rural environment. Experimental results show that our satellite-imagery-based localization achieves an average position error of 9.92m and an average heading error of $0.32^{\circ}$ over a 5km route in the urban environment, and an average position error of 9.29m and an average heading error of $0.42^{\circ}$ in the rural environment. \figref{fig:sat_map_loc} visualizes our satellite-imagery-based localisation in an urban environment. The top left image shows the ground-view panoramic image. The bottom left image shows the paths estimated by our localization and GNSS/INS system in green and red respectively on the bottom left. In the right image, particles are shown in blue on the right and interposed against a likelihood map; the more red the pixel, the higher the likelihood that the vehicle is located at that pixel.

\subsection{Sparse 3D Map Reconstruction}
\label{sec:map_reconstruction}

Sparse 3D map reconstruction is required for map-based 6-DoF localization which is described in \secref{sec:map_localization}. Prior to localization, we build a sparse 3D map in which each 3D point is associated with one or more local SIFT features \citep{Lowe2004IJCV}.

\begin{figure}
  \centering
	\subfloat[Bird's eye view.
		\label{fig:map_south_bv}]{%
		\includegraphics[width=0.9\columnwidth]{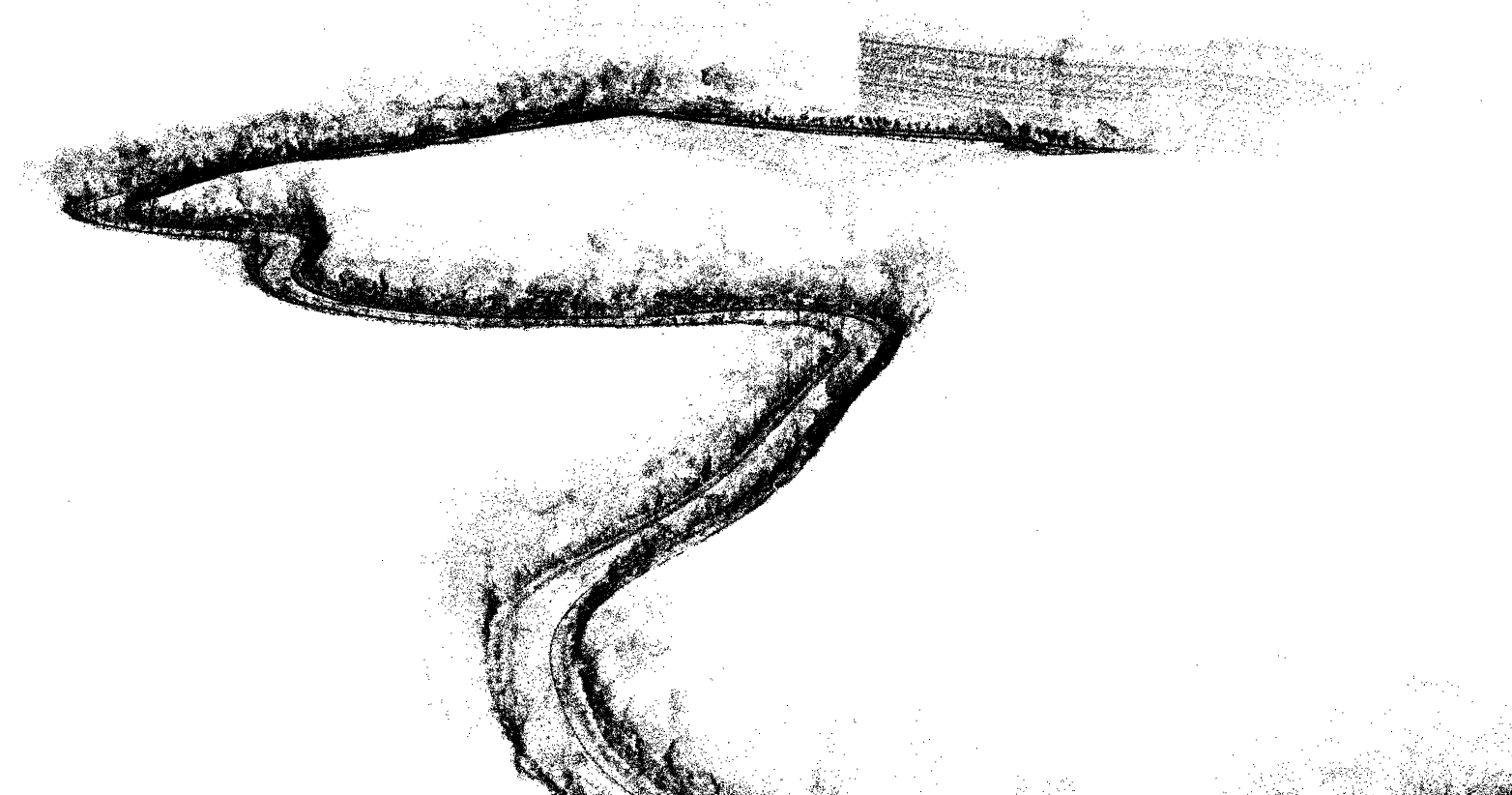}
     }
     \hfill
     \subfloat[Close-up view.
     	\label{fig:map_south_bv_close}]{%
		\includegraphics[width=0.9\columnwidth,trim={0 8cm 0 2cm},clip]{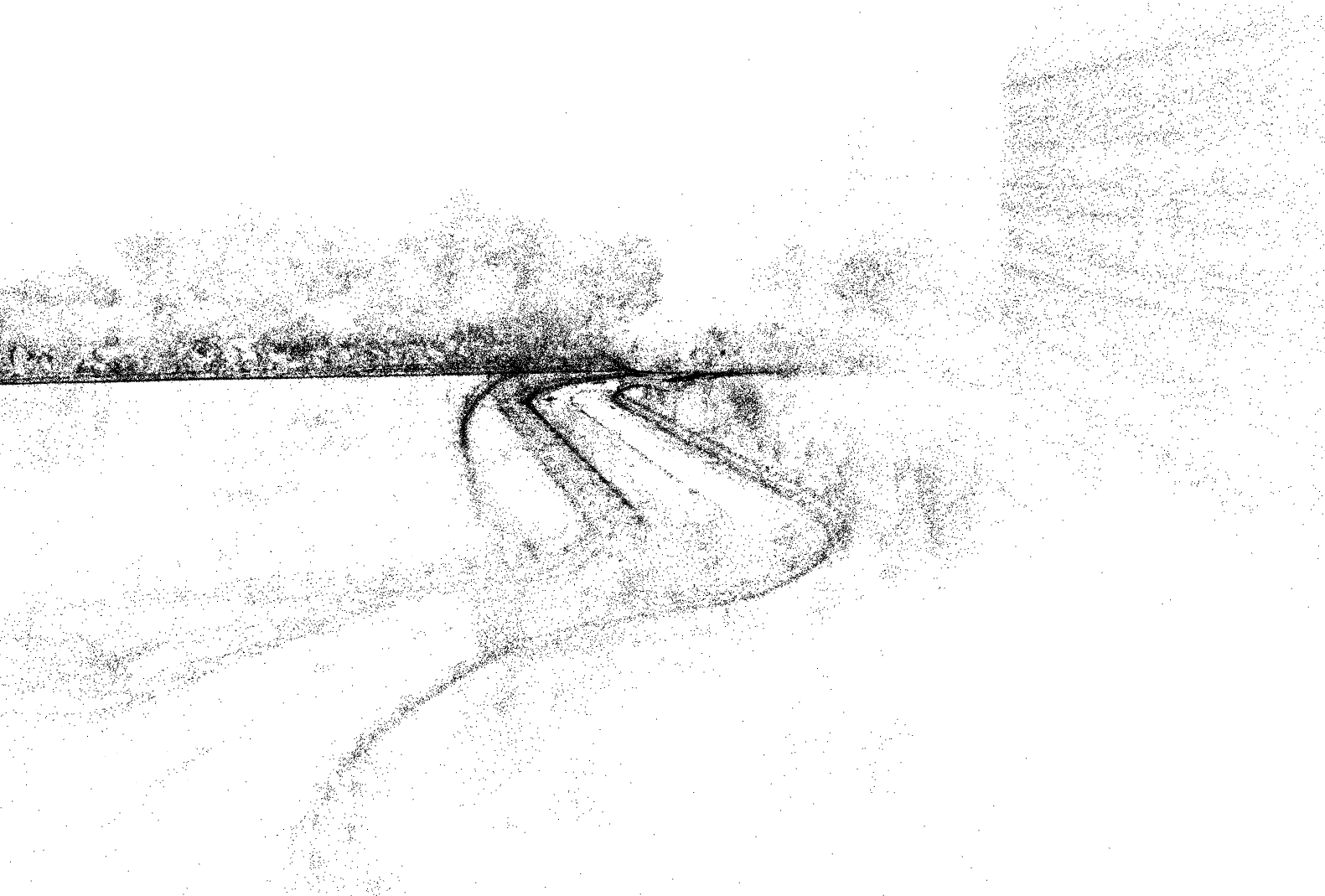}
     }
     \caption{The sparse 3D map of a mixed urban and rural environment and generated by our reconstruction pipeline.}
     \label{fig:reconstruction}
\end{figure}

To minimize the time required for large-scale reconstruction, our approach does not reconstruct the scene from scratch, and instead, uses reasonably accurate initial pose estimates from a GNSS/INS system to initialize all camera poses.
In addition, we require a minimum amount of camera motion between images used for mapping.
Next, we perform feature matching between nearby images, and use the feature matches and initial poses to triangulate the scene.
We then repeatedly optimize the scene structure and camera poses using bundle adjustment followed by the merging of feature tracks. 
This approach is implemented on top of the COLMAP structure-from-motion (SfM) framework \citep{Schoenberger2016CVPR}. 
During bundle adjustment, we enforce that the extrinsic parameters of the multi-camera system on the AutoVision vehicle remain constant.
\figref{fig:reconstruction} shows a sparse 3D map of a mixed urban and rural environment.

\subsection{6-DoF Localization with a Sparse 3D Map}
\label{sec:map_localization}

Given a sparse 3D map computed using the approach described
in \secref{sec:map_reconstruction} and without prior knowledge of the AutoVision vehicle's global pose, we localize the AutoVision vehicle by extracting local
SIFT features \citep{Lowe2004IJCV} from the images captured
by the cameras mounted on the AutoVision vehicle and matching
the descriptors of these 2D features against the descriptors
associated with the 3D points in the map. From the 2D-3D matches, we apply a generalized perspective-n-point pose solver \citep{Kneip2013ICRA, Lee2015IJRR} inside a RANSAC loop \citep{Fischler1981ACM} to estimate the vehicle's pose. With a sparse 3D map of a large area which contains many 3D points, 2D-3D matching is the main computational bottleneck in our pipeline. In this case, we use a prioritized matching approach based on Active Search \citep{Sattler2017PAMI} for improved matching efficiency. \citet{Geppert2019ICRA} describe our localization approach in greater detail. 

\begin{figure}
	\subfloat[Urban environment. In some parts of the area, the localization fails consistently to estimate a pose.\label{fig:map_loc_one_north_trajectory}]{%
		\includegraphics[width=0.48\columnwidth]{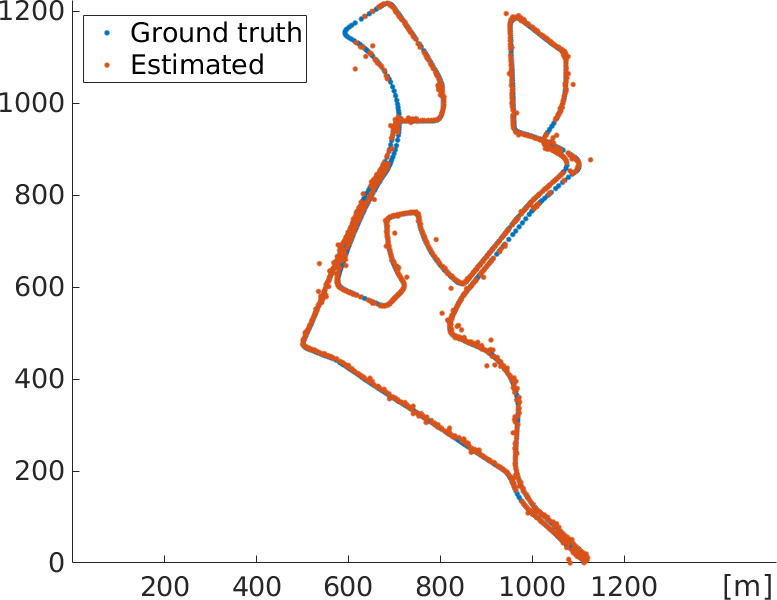}
     }
     \hfill
     \subfloat[Rural environment. The errors at the top end are likely caused by repetitive structures on both sides of the road.\label{fig:map_loc_south_bv_trajectory}]{%
		\includegraphics[width=0.48\columnwidth]{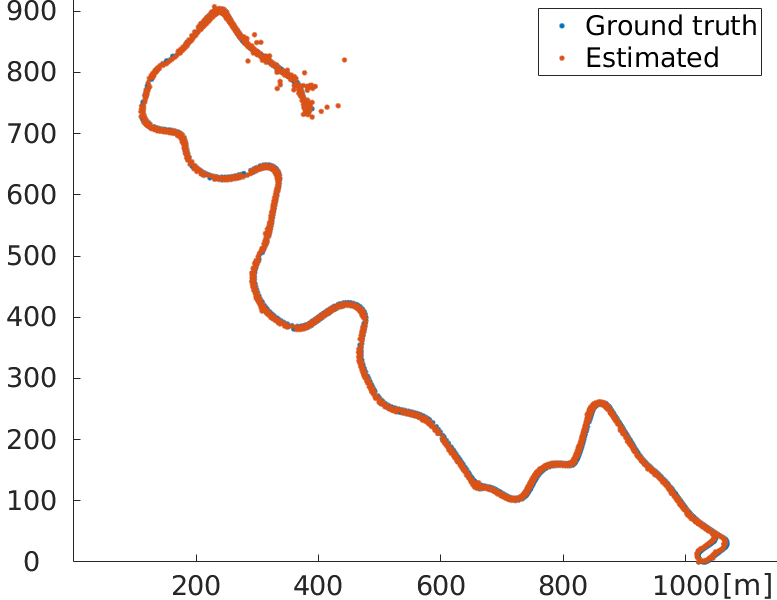}
     }
     \caption{The positions estimated by map-based localization vs. ground truth positions for a 5km route in two different environments.}
     \label{fig:map_localization}
\end{figure}

For our experiments, we use the same routes used for the satellite-imagery-based localization experiments described in \secref{sec:sat_map_localization}. Our localization pipeline runs at around 2 Hz on the AutoVision vehicle. \figref{fig:map_localization} shows the estimated and ground truth positions for the urban and rural routes. For the urban route, the mean and median errors of all reported poses are 3.31m and 1.84m for the position, and 2.6$^{\circ}$ and 1.9$^{\circ}$ for the heading, respectively. For the rural route, the mean and median errors of all reported poses are 3.48m and 1.81m for the position, and 4.2$^{\circ}$ and 3.3$^{\circ}$ for the heading, respectively.

\section{Perception}

\begin{figure}
	\subfloat[Image from the front center camera.\label{fig:dense_mapping_image}]{%
		\includegraphics[width=0.48\columnwidth]{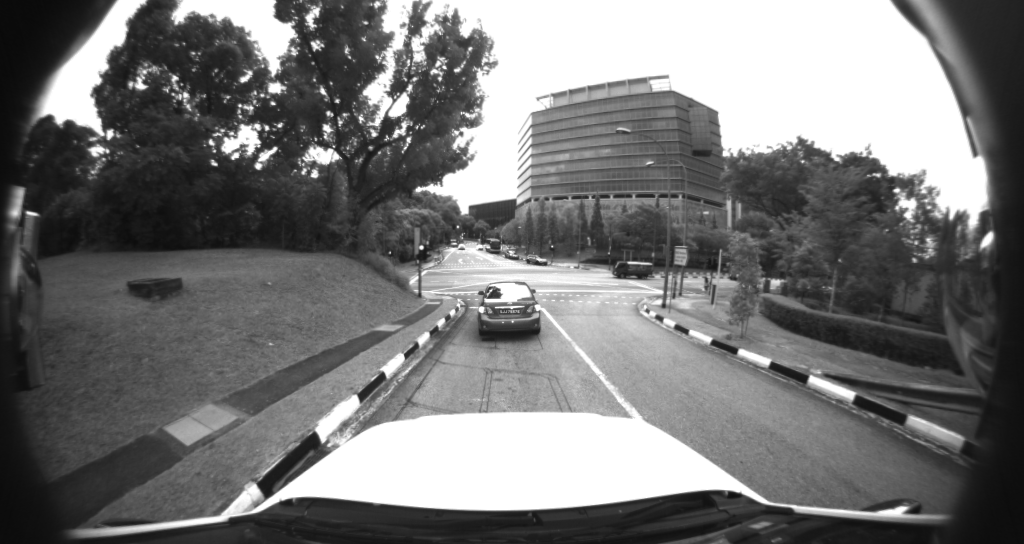}
     }
     \hfill
     \subfloat[Depth image generated by plane-sweeping stereo.\label{fig:dense_mapping_depth_image}]{%
		\includegraphics[width=0.48\columnwidth]{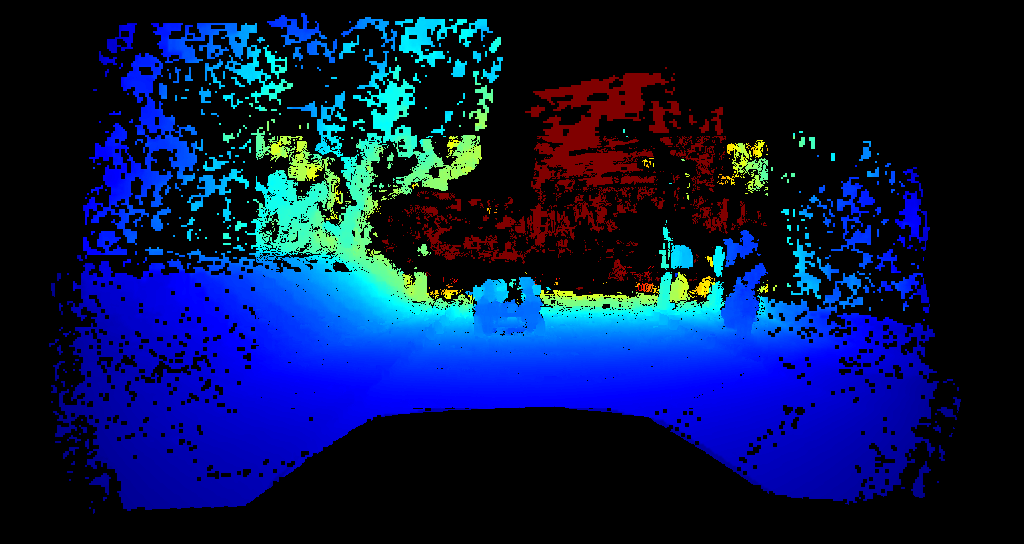}
     }
     \hfill
     \subfloat[Dynamic objects detected by the object detector.\label{fig:dense_mapping_detection}]{%
		\includegraphics[width=0.48\columnwidth]{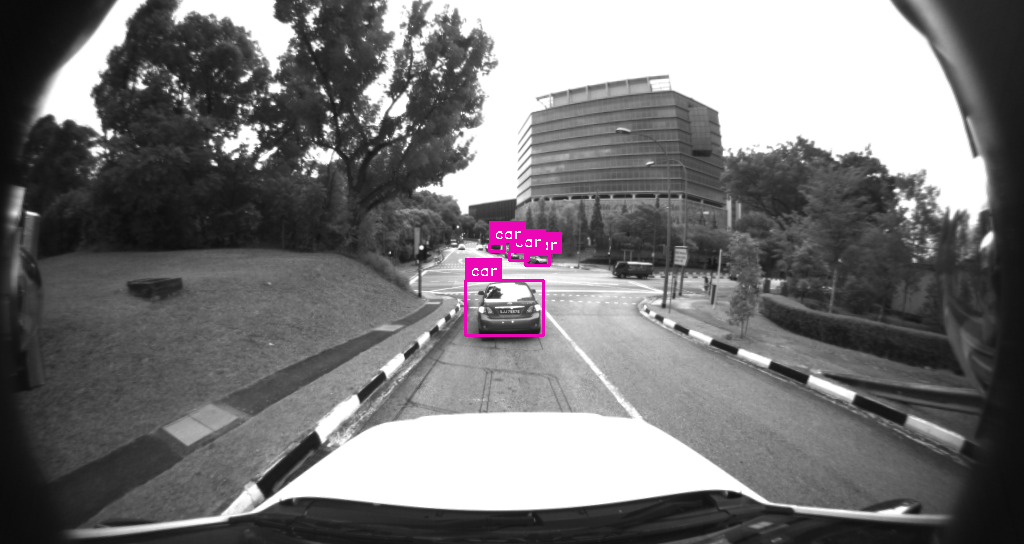}
     }
     \hfill
     \subfloat[Raycasted TSDF volume.\label{fig:dense_mapping_tsdf}]{%
		\includegraphics[width=0.48\columnwidth]{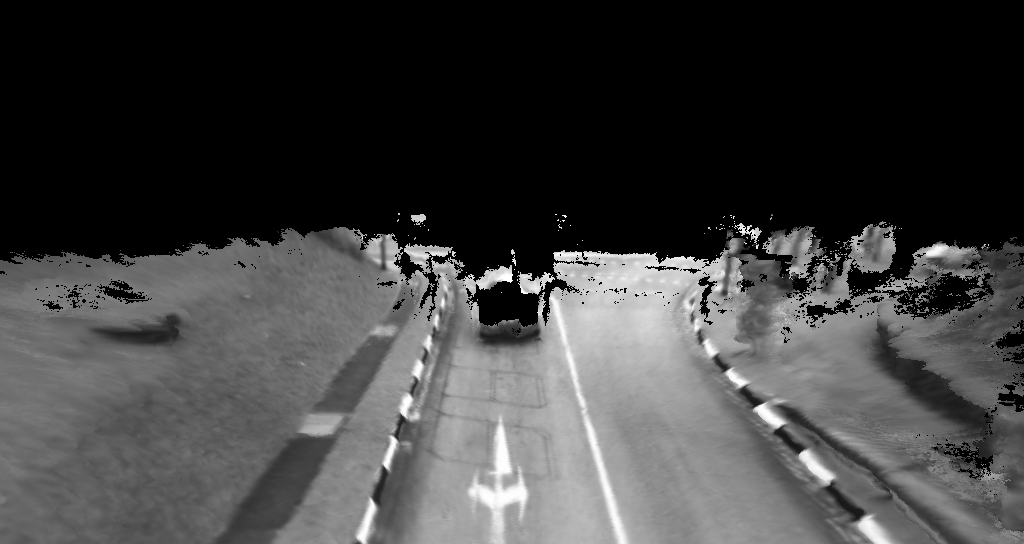}
     }
     \caption{Output from the AutoVision perception modules \citep{Cui2019ICRA}.}
     \label{fig:dense_mapping}
\end{figure}

Dense 3D mapping in real-time is a prerequisite for autonomous navigation in complex environments. Our perception pipeline begins with plane-sweeping stereo generating depth images. In turn, depth images are fused into a truncated signed distance function (TSDF) volume. A 3D map is reconstructed from this TSDF volume via ray-casting. To avoid dynamic objects from corrupting the 3D map via trails of artefacts, we detect potentially dynamic objects not belonging to the static environment, and remove their associated depth estimates from the depth images prior to depth fusion. \figref{fig:dense_mapping} visualizes the outputs from plane-sweeping stereo, object detection, and TSDF-based 3D mapping. More details of our perception pipeline can be found in \citep{Cui2019ICRA}.

\subsection{Plane-Sweeping Stereo}

Plane-sweeping stereo computes a depth image for multiple images with known
camera poses by sweeping a set of planes through 3D space.
Each plane represents a depth hypothesis and defines a homography
mapping from every other view to the reference view.
We estimate the depth for each pixel in a reference image by
using each plane to warp each non-reference image to the reference image,
evaluating the image dissimilarity at that pixel, and choosing
the plane that minimizes the image dissimilarity.
We use the GPU implementation of plane-sweeping stereo for fisheye
cameras \citep{Haene20153DV} which computes depth images directly
from fisheye images without the need for undistortion, allowing us to use
the full field-of-view of the cameras. On the AutoVision vehicle, plane-sweeping stereo runs at an
average of 15 Hz for the 5 cameras at the front of the vehicle and with images
downsampled to half-resolution.

\subsection{TSDF-based 3D Mapping}

A single depth image may not contain sufficient geometric information for up-stream modules such as 3D semantic segmentation and motion planning. Hence, we need to fuse depth images estimated at different positions in time to create a dense 3D map. We use a standard fusion technique: the scene is represented via a set of voxels where each voxel stores a TSDF value \citep{Curless1996SIGGRAPH}.
Here, each voxel stores the signed distance to the closest
object surface (negative inside of objects, positive outside
of objects, zero on surfaces), truncated to a certain maximum
/ minimum value. Whenever a new depth image along
with its camera pose becomes available, we update the 3D
model. We use the map fusion pipeline in the InfiniTAM
library \citep{Kahler2015ISMAR, Kahler2016RAL}.
We also use the fast raycasting algorithm in \citep{Kahler2015ISMAR, Kahler2016RAL} to reconstruct the 3D map in the current camera view.
The pipeline runs at around 20 Hz on the AutoVision vehicle.

\subsection{Dynamic Object Detection}

Dynamic objects leave behind trails of artefacts in the 3D map. We leverage 2D object detection to solve this problem. Given a reference image, we detect dynamic objects, i.e. humans and vehicles. In turn, for the corresponding depth image, we mask out pixels located within the 2D bounding boxes of detected objects. This way, we avoid integrating depth estimates associated with dynamic objects into the 3D map. We use the YOLOv3 object detection network \citep{Redmon2018CoRR} trained on the Microsoft COCO dataset \citep{Lin2014ECCV}. To improve the inference performance with distorted grayscale images from our NIR fisheye cameras, we fine-tune the network by truncating the first and last layers, and retrain the network using our labeled datasets.

\section{CONCLUSIONS}

Project AutoVision has successfully demonstrated localization and 3D scene perception for autonomous vehicles with multi-camera systems, in both urban and rural environments, and without GNSS. As Project AutoVision progresses, we will continue to enhance localization and perception capabilities, and add more modules to our software stack. These modules include but are not limited to, change detection for 3D maps, obstacle detection, dynamic object tracking and classification, and semantic 3D mapping.


\bibliographystyle{abbrvnat}
\bibliography{references}

\end{document}